%% file: acl_latex.tex
\pdfoutput=1

\documentclass[11pt]{article}

\usepackage[preprint]{acl}

\usepackage{times}
\usepackage{latexsym}

\usepackage[T1]{fontenc}

\usepackage[utf8]{inputenc}

\usepackage{microtype}

\usepackage{inconsolata}

\usepackage{graphicx}
\usepackage{balance}

\usepackage{amsmath}
\usepackage{amssymb}
\usepackage{enumitem}
\usepackage{cleveref}
\usepackage{wrapfig}

\crefname{figure}{Fig.}{Figs.}
\crefname{table}{Tab.}{Tabs.}
\crefname{section}{Sec.}{Secs.}
\crefname{subsection}{Subsec.}{Subsecs.}
\crefname{equation}{Eq.}{Eqs.}
\crefname{theorem}{Thm.}{Thms.}
\crefname{lemma}{Lem.}{Lems.}
\crefname{algorithm}{Alg.}{Algs.}

\usepackage{booktabs}

\title{Trajectory Prediction Meets Large Language Models: A Survey}

\author{Yi Xu \quad Ruining Yang \quad Yitian Zhang \quad Jianglin Lu\\
    \bf Mingyuan Zhang  \quad Yizhou Wang\quad Lili Su \quad Yun Fu \\
    Department of Electrical and Computer Engineering, Northeastern University\\
    \texttt{\{xu.yi,yang.ruini,lu.jiang,zhang.mingyua,l.su\}@northeastern.edu}\\
    \texttt{\{markcheung9248,wyzjack990122\}@gmail.com}\\
    \texttt{yunfu@ece.neu.edu}
    }

\begin{document}
\maketitle

\input{section/0_abstract}

\noindent
\begin{wrapfigure}{l}{0.05\textwidth}
\vspace{-4mm}
    \centering
    \href{https://github.com/colorfulfuture/Awesome-Trajectory-Motion-Prediction-Papers}{\includegraphics[width=0.05\textwidth]{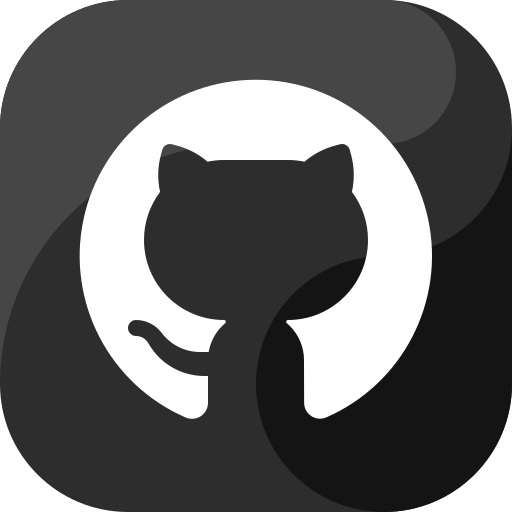}}
\end{wrapfigure}
{\fontsize{8}{8}\selectfont
  \href{https://github.com/colorfulfuture/Awesome-Trajectory-Motion-Prediction-Papers}{%
    \texttt{https://github.com/colorfulfuture/}\\%
    \texttt{Awesome-Trajectory-Motion-Prediction-Papers}%
  }%
}\\

\input{section/1_introduction}
\input{section/2_background}
\input{section/3_taxonomy}
\input{section/4_method}
\input{section/5_challenges}
\input{section/7_conclusion}

\input{section/limitations}

\input{section/ethics}


\bibliography{custom}

\appendix
\input{section/x_appendix}

\end{document}

%% file: section/0_abstract.tex
\begin{abstract}
Recent advances in large language models (LLMs) have sparked growing interest in integrating language-driven techniques into trajectory prediction. 
By leveraging their semantic and reasoning capabilities, LLMs are reshaping how autonomous systems perceive, model, and predict trajectories.
This survey provides a comprehensive overview of this emerging field, categorizing recent work into five directions: 
(1) Trajectory prediction via language modeling paradigms, 
(2) Direct trajectory prediction with language-based predictors, 
(3) Language-guided scene understanding for trajectory prediction,
(4) Language-driven data generation for trajectory prediction, 
(5) Language-based reasoning and interpretability for trajectory prediction.
For each, we analyze representative methods, highlight core design choices, and identify open challenges. This survey bridges natural language processing and trajectory prediction, offering a unified perspective on how language can enrich trajectory prediction.
\end{abstract}

%% file: section/1_introduction.tex
\section{Introduction}
Trajectory prediction is a critical task in autonomous systems, with applications in autonomous driving~\citep{huang2022survey, madjid2025trajectory}, robot navigation~\citep{rosmann2017online, bhaskara2024trajectory}, interaction modeling~\citep{huang2019stgat, zhu2021simultaneous}, and multi-agent coordination~\citep{zhao2019multi, xu2025sports}. 
Traditional approaches rely on geometric or physics-based formulations or patterns extracted from past data. 
A core challenge lies in accurately modeling the interactions among agents.
Although recent data-driven models learn these dependencies from large datasets, their generalization still hinges on data diversity and quality.

Large language models (LLMs)~\citep{zhao2023survey, chang2024survey} and multimodal reasoning~\citep{wei2022chain, plaat2024reasoning} open new avenues for connecting high-level semantics with low-level motion prediction. 
Language naturally conveys context, intent, and causality, allowing descriptions of scenes, goals, and potential futures. 
This expressiveness has driven growing interest in integrating LLMs into trajectory prediction, whether for contextual reasoning, reformulating trajectories as language sequences, or enabling data generation and interpretability.

\begin{figure}[t]
  \includegraphics[width=\columnwidth]{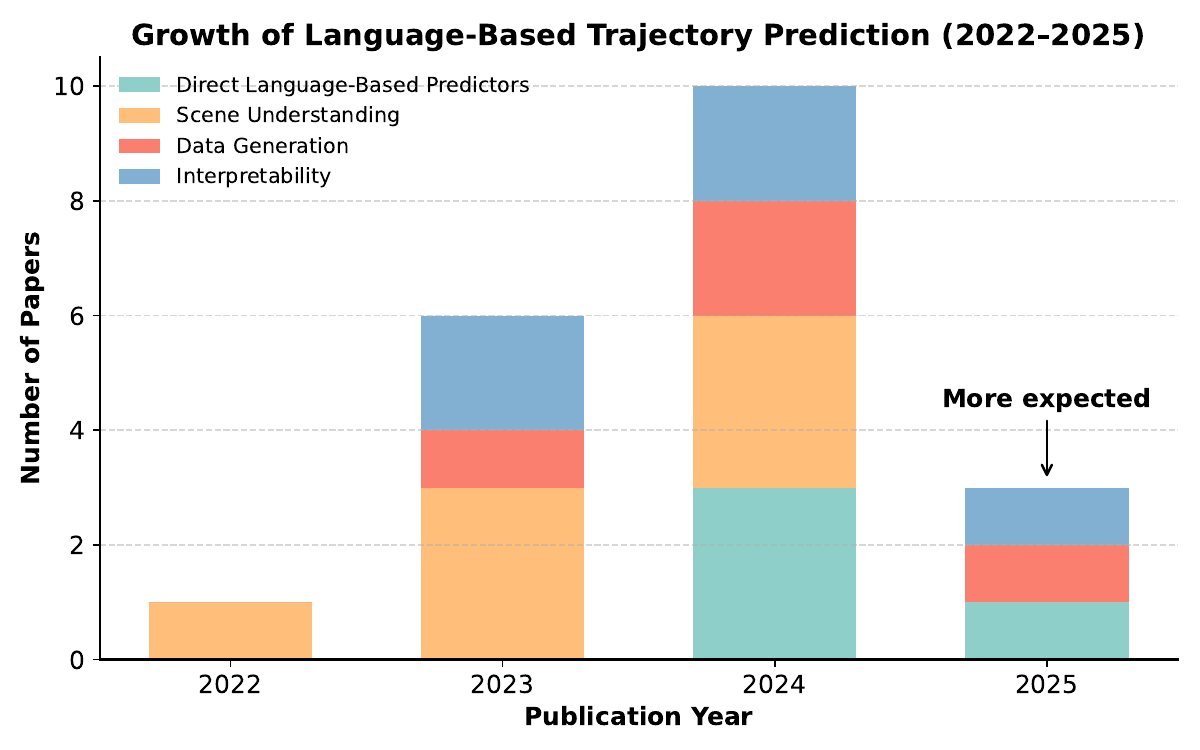}
  \caption{Growth of language-based trajectory prediction papers (2022–2025) across four categories. Only methods explicitly incorporating natural language or pretrained language models (PLMs/LLMs) are counted.}
  \label{fig:teaser}
\end{figure}

\begin{figure*}[t!]
  \includegraphics[width=\linewidth]{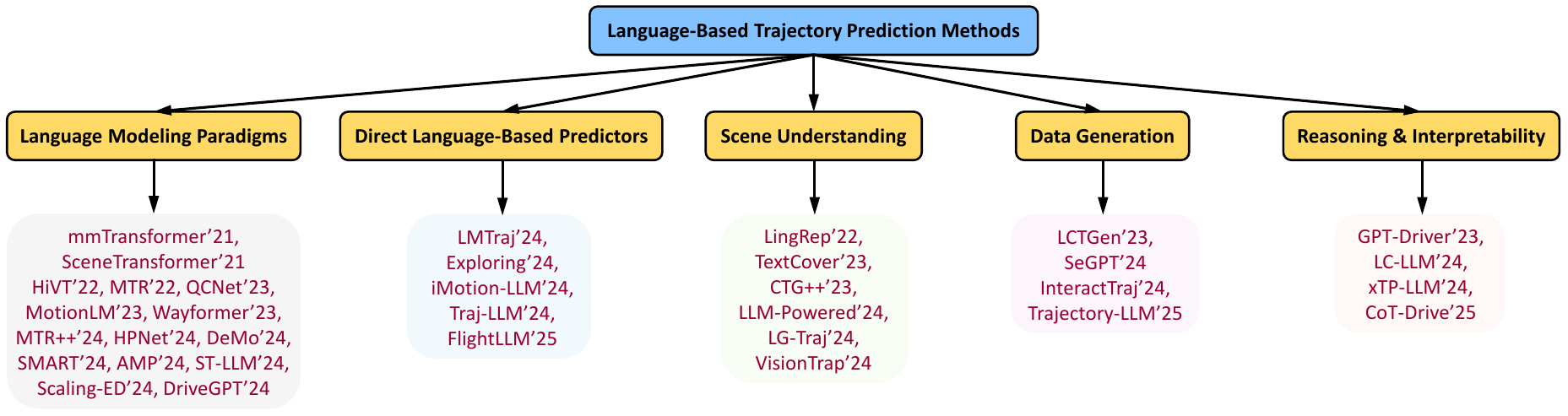} 
  \caption{Taxonomy of language-based trajectory prediction methods. We categorize representative works (2022–2025) into five groups based on the functional role of language: \textit{language modeling paradigms}, \textit{direct language-based prediction}, \textit{scene understanding}, \textit{data generation}, and \textit{reasoning \& interpretability}.}
  \label{fig:overview}
\end{figure*}

This survey provides the first systematic overview of how LLMs are reshaping trajectory prediction. 
The field, though young, is expanding rapidly (Figure~\ref{fig:teaser}), underscoring the need for a unified synthesis. 
We summarize existing studies, identify modeling patterns, and discuss open challenges to guide future research.
Our contributions are summarized as follows:
\begin{itemize}[leftmargin=*, labelsep=0.5em, itemindent=0pt]
    \item To the best of our knowledge, this is the first survey on the growing intersection of LLMs and trajectory prediction, bridging perspectives from both NLP and trajectory prediction communities.
    \item We systematically organize existing work into five categories and analyze each in terms of modeling paradigms, architecture, and the role of language across different prediction stages.
    \item We discuss open challenges and outline future directions aimed at better aligning language understanding with trajectory prediction objectives.
\end{itemize}


%% file: section/2_background.tex
\section{Background}\label{sec:background}
\subsection{Trajectory Prediction Formulation}
Trajectory prediction forecasts the future positions of agents from their observed states and scene context. 
Given observations $\mathbf{X} \in \mathbb{R}^{N \times H \times F}$ for $N$ agents over $H$ steps, the task is to predict $\mathbf{Y} \in \mathbb{R}^{N \times T \times F}$ over a horizon $T$, where $F$ is typically 2 for 2D coordinates. 
Additional context $\mathbf{I}$ (e.g., maps or layouts) is often included, and models usually learn a probabilistic distribution $p_\theta(\mathbf{Y}\mid \mathbf{X},\mathbf{I})$ to capture multiple plausible futures.

The problem remains challenging due to multimodal uncertainty, long-range temporal dependencies, and heterogeneous inputs. 
Recent studies explore LLMs as a means to address these issues.

\subsection{Key Concepts in Large Language Models}
Large Language Models (LLMs) are Transformer-based networks~\citep{vaswani2017attention} trained on internet-scale data, excelling at language understanding and reasoning. 
Their autoregressive design models sequential dependencies, and pretraining enables zero- or few-shot generalization~\citep{wei2021finetuned, kojima2022large}. 
These traits align naturally with trajectory prediction, which requires semantic abstraction and intent reasoning among interacting agents.

To leverage LLMs for trajectory tasks, spatial or contextual data are converted into token sequences or descriptive prompts. 
Some works directly employ pretrained models such as GPT~\citep{floridi2020gpt, achiam2023gpt}, while others adapt them via fine-tuning~\citep{ziegler2019fine}, instruction tuning~\citep{zhang2023instruction}, or lightweight adapters (e.g., LoRA~\citep{hu2022lora}). 
These adaptations enable LLMs to interpret structured inputs, generate diverse trajectories, and produce linguistic rationales within a unified framework.

%% file: section/3_taxonomy.tex
\section{Taxonomy}\label{sec:taxonomy}
We categorize the literature into five distinct yet interconnected directions, as illustrated in Figure~\ref{fig:overview}. 
These categories capture the methodological diversity across recent studies and highlight the different roles that language plays in enhancing predictive capability. 
The survey methodology, including search criteria and selection pipeline, is described in Appendix~\ref{sec:appendix_meth}.
\begin{itemize}[leftmargin=*, labelsep=0.5em, itemindent=0pt]
    \item \textbf{Language Modeling Paradigms:} Trajectories are tokenized and modeled autoregressively, enabling long-range temporal reasoning.
    \item \textbf{Direct Language-Based Predictors:} Pretrained LLMs are directly used via prompting or fine-tuning for trajectory generation.
    \item \textbf{Scene Understanding:} Language augments spatial context by encoding semantics such as traffic rules or agent intent.
    \item \textbf{Data Generation:} LLMs synthesize diverse and controllable trajectory datasets from text.
    \item \textbf{Reasoning \& Interpretability:} Models generate natural-language rationales to explain or justify their predictions.
\end{itemize}

\input{table/summary_main}

We further provide a quantitative summary of representative works in Table~\ref{tab:summary_main}, highlighting datasets, metrics, and evaluation conventions across the five categories.
The appendix (Appendix~\ref{sec:appendix_C}) expands this summary with detailed per-category comparisons and model configurations.

%% file: table/summary_main.tex
\begin{table*}[t]
\centering
\scalebox{0.75}{
\begin{tabular}{lccc}
\toprule
\textbf{Category} & \textbf{Representative Methods} & \textbf{Common Datasets} & \textbf{Typical Metrics} \\
\midrule
Language Modeling Paradigms & mmTransformer, MTR, MotionLM & Argoverse, WOMD & ADE / FDE / MR \\

Direct Language-Based Predictors & LMTraj, iMotion-LLM, Traj-LLM & WOMD, nuScenes & ADE / FDE \\

Language-Guided Scene Understanding & CTG++, LLM-Powered, VisionTrap & nuScenes, BDD100K & ADE / FDE / Qualitative \\

Language-Driven Data Generation & LCTGen, InteractTraj, Trajectory-LLM & L2T, nuScenes-Sim & – (Synthetic) \\

Reasoning and Interpretability & LC-LLM, CoT-Drive, GPT-Driver & WOMD, xTP & ADE / FDE + Text Eval \\
\bottomrule
\end{tabular}
}
\caption{Summary-level comparison across representative works.}
\label{tab:summary_main}
\end{table*}

%% file: section/4_method.tex
\section{Trajectory Prediction via Language Modeling Paradigms}\label{sec:taxonomy_1}
Trajectory prediction has been strongly shaped by advances in NLP~\citep{vaswani2017attention, devlin2019bert}, particularly Transformer architectures and attention mechanisms.
Originally designed for language sequences, Transformers effectively capture long-range dependencies and multi-agent interactions. 
Recent works thus adapt these sequence modeling paradigms for representing and generating trajectories, reflecting an ongoing convergence between NLP and motion forecasting.

One early contribution is mmTransformer~\citep{liu2021multimodal}, which employs a stacked Transformer framework for multimodal motion prediction. It follows a two-stage design that first generates diverse trajectory candidates and then refines them through attention-based decoding. A region-based training strategy encourages coverage over multiple plausible futures, laying the groundwork for applying attention to multimodal trajectory modeling.
Another representative work, SceneTransformer~\citep{ngiam2021scene}, introduces a unified Transformer model that jointly predicts multiple agent trajectories, using a masking strategy inspired by NLP to flexibly condition on goals. By integrating information across road elements and agent interactions, it demonstrates the strength of attention for capturing agent dependencies, though at high computational cost and with limited semantic grounding.
Collectively, these early architectures demonstrate the promise of attention for modeling interactions, while revealing two key limitations, computational inefficiency and lack of explicit semantics, that later works address through hierarchical and geometry-aware modeling.


HiVT~\citep{zhou2022hivt} introduces a hierarchical vectorized Transformer that separates local and global interactions through structured attention, improving representational efficiency and geometric consistency.
Wayformer~\citep{nayakanti2023wayformer} further unifies scene encoding and trajectory decoding via modular fusion and latent query attention, exploring early, late, and hierarchical integration of input modalities such as maps, agents, and signals.
Together, these architectures emphasize structured fusion and query-based attention as effective biases for interaction modeling, though they still lack explicit semantic grounding from language.


Building on these structured fusion designs, Motion Transformer (MTR)~\citep{shi2022motion} introduces a two-branch framework that decouples global intention localization from local motion refinement.
Instead of enumerating dense trajectory candidates, it employs learnable motion queries, each representing a distinct future mode.
This query-based formulation improves diversity and training stability, achieving strong results on the Waymo Open Motion Dataset (WOMD)~\citep{ettinger2021large}.
MTR++~\citep{shi2024mtr++} extends this idea to multi-agent settings through symmetric scene modeling and mutually guided intention queries, enabling agents to coordinate predictions via structured attention and generate scene-consistent trajectories.
These intention-query paradigms advance multimodal coverage and interaction reasoning, though they remain sensitive to mode granularity and candidate sampling.


Building on intention-query designs, QCNet~\citep{zhou2023query} shifts the focus to an agent-centric query formulation, where each target agent retrieves context most relevant to its own motion.
This selective querying improves both prediction accuracy and interpretability, particularly in dense traffic scenarios.
Complementing this direction, HPNet~\citep{tang2024hpnet} enhances temporal reasoning by feeding past predictions back into the attention mechanism.
Its historical prediction attention explicitly links prior forecasts with future motion, improving stability over long horizons.
Extending disentangled modeling further, DeMo~\citep{zhang2024demo} separates trajectory generation into two complementary query types, directional mode queries to encode intent and dynamic state queries to model motion evolution.
By integrating attention with state-space modeling~\citep{gu2023mamba}, it achieves fine-grained control over trajectory diversity and physical plausibility.
Together, these works demonstrate a progressive refinement from query-centric interaction modeling toward modular reasoning frameworks that unify attention and dynamics.


Building on Transformer-based architectures, recent research has moved beyond structural borrowing to directly adopt language modeling paradigms for trajectory generation.
The rise of LLMs motivates viewing trajectory prediction as an autoregressive token-generation task, where continuous motion is discretized into symbolic tokens and generated step by step, much like words in a sentence.
This shift marks a conceptual transition from attention-based interaction modeling to full language-style sequence generation.


MotionLM~\citep{seff2023motionlm} first formulated motion forecasting as a language modeling problem, using a causal Transformer decoder to jointly predict multi-agent futures.
SMART~\citep{wu2024smart} extended this to real-time simulation, achieving low-latency inference through tokenized map-trajectory representations.
Building on these ideas, AMP~\citep{jia2024amp} unified token space design and factorized attention to enhance structure-aware prediction, while ST-LLM~\citep{liu2024spatial} further integrated spatial–temporal tokenization for large-scale forecasting.
Together, these models show how discrete motion tokens enable unified sequence reasoning and improved long-horizon consistency.

Despite strong accuracy, the computational cost of autoregressive decoding remains high.
Scaling-ED~\citep{ettinger2024scaling} addresses this by distilling large ensembles into compact student models, achieving a practical balance between performance and efficiency.
Pushing scale further, DriveGPT~\citep{huang2024drivegpt} applies autoregressive Transformers to billions of driving frames, empirically validating scaling laws and bridging trajectory prediction with foundation model trends.

These works indicate the shift from architecture transfer to paradigm transfer, from using attention as a tool to adopting language modeling as a principle.
While scaling and tokenization improve generalization, challenges remain in spatial grounding and physical realism, highlighting the next frontier for language-inspired prediction systems.



\noindent\textbf{Discussion.}
Language-modeling paradigms have brought new structure and abstraction to trajectory prediction, introducing tokenized motion generation, scalable sequence reasoning, and cross-agent attention mechanisms.
Yet, most remain architectural transfers from NLP rather than semantically grounded systems: they excel at temporal coherence but struggle with spatial precision and interpretability once trajectories are discretized.
A core tension emerges between scalability and physical realism.
Large autoregressive models generalize well with data scale but often overlook continuous dynamics and agent intent.
Furthermore, evaluation protocols still center on geometric error (e.g., ADE/FDE) rather than semantic fidelity, leaving reasoning quality underexplored.
Future directions may include hybrid formulations that integrate discrete sequence modeling with continuous spatial reasoning, enabling models to move beyond syntactic prediction toward genuine semantic understanding of motion and intent.


\section{Direct Trajectory Prediction with Language-Based Predictors}\label{sec:taxonomy_2}
Beyond cataloging representative systems, this section examines how language conditioning aids trajectory prediction, in intent control, few-shot transfer, and cross-domain generalization, and where it still falls short, such as numerical precision and spatial grounding.
Unlike Section~\ref{sec:taxonomy_1}, which focused on architectures inspired by NLP, the methods here employ pretrained language models (PLMs/LLMs) directly, either fine-tuned or used in zero-/few-shot modes, as the predictive core for trajectory generation.
This shift reflects a move from architectural borrowing to language models as active predictors, leveraging their reasoning and generalization strengths across domains and agent types.
In this section, we refer to both fine-tuned pretrained language models (PLMs) such as T5 and large-scale generative models like GPT-3.5/4 as language-based predictors.
While they differ in scale, both function as pretrained, language-conditioned modules within trajectory pipelines.


LMTraj~\citep{bae2024can} reformulates trajectory prediction as a question-answering task, converting numerical and scene features into structured prompts.
Through a decimal-aware tokenizer and auxiliary interaction objectives, it bridges the gap between continuous values and discrete tokens, outperforming regression baselines.
This shows that linguistic prompting can encode high-level reasoning, though it remains sensitive to token granularity.
Expanding this direction, \citet{munir2024exploring} benchmarks multiple open-source LLMs, including GPT-2~\citep{radford2019language}, LLaMA-7B~\citep{touvron2023llama}, LLaMA-7B-Chat, Zephyr-7B~\citep{tunstall2023zephyr}, and Mistral-7B~\citep{jiang2023mistral}, on the trajectory prediction task.
They find strong intent reasoning but weak numerical grounding, underscoring the need for domain-specific adaptation.
To strengthen intent alignment, iMotion-LLM~\citep{felemban2024imotion} applies instruction tuning via LoRA fine-tuning~\citep{hu2022lora} on WOMD, pairing language directives (e.g., ``turn left at the next intersection'') with motion data.
This demonstrates language as a flexible control interface, though template-based commands may limit generalization.
FlightLLM~\citep{luo2025large} extends this paradigm to aviation, serializing flight paths into tokens and fine-tuning LLaMA for short- and long-term forecasting, while noting inference latency as a key barrier.
Finally, Traj-LLM~\citep{lan2024traj} integrates LLMs directly into the trajectory pipeline, encoding multimodal scene context into transformer-compatible sequences.
Using a GPT-2 backbone with LoRA fine-tuning and a lane-aware probabilistic head, it maintains high accuracy even with limited data, highlighting LLMs’ few-shot adaptability.

Together, these methods illustrate the transition from using language as inspiration to using it as computation, LLMs serving as trainable predictors.
Despite clear gains in interpretability and cross-domain transfer, persistent issues remain in grounding, precision, and efficiency, calling for hybrid or hierarchical designs that reconcile reasoning power with spatial fidelity.



\noindent\textbf{Discussion.}
Direct language-based predictors embed reasoning directly into motion forecasting, excelling in few-shot generalization and intent conditioning.
Yet their reliance on tokenization and autoregressive decoding leads to numerical imprecision and latency bottlenecks, complicating deployment.
Most current systems trade spatial control for interpretability: they can explain intent but struggle to anchor it geometrically.
A promising path lies in hybrid models that couple continuous spatial embeddings with linguistic prompts, or multi-stage fine-tuning schemes that retain reasoning while improving precision and efficiency.


\section{Language-Guided Scene Understanding for Trajectory Prediction}\label{sec:taxonomy_3}
Traditional trajectory predictors mainly rely on structured inputs, agent coordinates, map layouts, or categorical signals, that often fail to capture scene semantics or intent-level interactions.
To bridge this gap, recent studies introduce natural language as a complementary modality, either through textual inputs, intermediate representations, or language-conditioned reasoning modules.
Language-guided cues enrich perception, enabling more interpretable and socially aware prediction in complex environments.


LingRep~\citep{kuo2022trajectory} first used template-based descriptions (e.g., ``crossing at intersection'', ``following another car'') as intermediate representations to interpret agent behavior.
While effective for interpretability, its rule-based design limits scalability and domain transfer.
Moving beyond handcrafted templates, TextCover~\citep{keysan2023can} integrates DistilBERT embeddings into a Transformer backbone, fusing textual and spatial representations for joint reasoning over context and motion.
Results show that pretrained textual embeddings can enhance urban-scene prediction, but only when spatial–text alignment is well learned.


Pushing this idea further, CTG++~\citep{zhong2023language} adopts diffusion-based generation guided by natural-language prompts.
GPT-4 converts free-form user descriptions (e.g., ``a group of pedestrians crossing while a car waits'') into differentiable control functions, enabling scenario-level synthesis.
Similarly, LLM-Powered~\citep{zheng2024large} embeds GPT-4V-derived semantic cues into the attention layers of a motion-prediction network, improving focus on socially relevant agents.
These efforts highlight the emerging role of multimodal LLMs as social reasoners, though their computational overhead and dependency on prompt quality remain barriers to deployment.


Focusing on pedestrian interactions, LG-Traj~\citep{chib2024lg} summarizes prior motion and visual context into a language-derived scene embedding, improving intent disambiguation in crowded settings.
VisionTrap~\citep{moon2024visiontrap} goes a step further by using vision–language models (VLMs) to auto-generate scene-level cues (e.g., ``child near crosswalk'') that guide attention toward salient regions.
Despite such gains, current VLM-based pipelines still face limited visual grounding and high inference cost.


Together, these works trace a progression from language as description to language as control signal.
By encoding scene semantics and relational context in natural language, they illustrate language’s growing value for context-aware and socially aligned prediction.
As alignment improves, language-conditioned models may become key to bridging perception, reasoning, and motion generation in complex environments.


\noindent\textbf{Discussion.}
Language-guided approaches show how linguistic cues can link low-level perception with high-level intent reasoning, enhancing interpretability and contextual awareness in dense or ambiguous scenes.
Yet, their success hinges on multimodal alignment quality, imbalanced training often leads to misplaced attention or incoherent goal inference.
The main trade-off lies between semantic richness and efficiency: stronger language integration yields better reasoning but higher inference cost.
Future work could explore lightweight adapters for linguistic grounding, joint spatial–text training, and benchmarks that quantify semantic alignment.
Reporting text-to-region grounding metrics (e.g., referring-expression accuracy) alongside ADE/FDE would make progress on reasoning and perception more measurable.


\section{Language-Driven Data Generation for Trajectory Prediction}\label{sec:taxonomy_4}
As trajectory predictors grow increasingly data-hungry and domain-specific, recent work explores LLMs as generative engines for creating synthetic data.
Natural language serves as a high-level interface to describe, simulate, and construct diverse trajectory scenarios, offering a scalable alternative to data collection and rule-based scene design.


LCTGen~\citep{tan2023language} first proposed using GPT-4~\citep{achiam2023gpt} as an interpreter to synthesize traffic scenes from natural-language prompts (e.g., ``a vehicle overtakes on the left while a pedestrian crosses'').
A Transformer-based decoder then generates agent distributions and motions grounded in map topology, establishing early feasibility for scene-level linguistic control, albeit with limited physical fidelity and diversity.
Building on this idea, SeGPT~\citep{li2024chatgpt} employs chain-of-thought~\citep{wei2022chain} prompting with ChatGPT to convert free-form text into structured driving configurations.
Although trajectories remain simplified, the framework illustrates how pretrained LLMs can act as controllable scenario generators for simulation and planning research, while also revealing the difficulty of enforcing physical plausibility from purely textual reasoning.


To enhance behavioral realism, InteractTraj~\citep{xia2024language} translates abstract descriptions into structured motion codes via GPT-4, later decoded into multi-agent trajectories capturing interactions such as yielding, merging, or aggressive acceleration.
This pipeline balances linguistic expressiveness with physical realism, though each translation stage introduces uncertainty.
Most recently, Trajectory-LLM~\citep{yang2025trajectory} adopts LLaMA-7B to convert compact textual descriptions of interactions into multimodal trajectory data.
Its three-stage Interaction–Action–Trajectory process, paired with map encoding, yields the L2T dataset of 240K text–map–trajectory samples, demonstrating a scalable route to semantically rich data generation.
However, such large-scale synthesis raises concerns about annotation quality, scenario balance, and cross-domain reproducibility.


Together, these works redefine language as not only a control interface but also a source of structured supervision.
By producing richly annotated and semantically coherent synthetic data, language-driven frameworks open new directions for cross-domain training, evaluation, and simulation in trajectory prediction.


\noindent\textbf{Discussion.}
Language-driven data generation offers a powerful yet underexplored frontier for trajectory modeling.
Using language as a generative interface simplifies scenario specification and broadens data diversity, but realism and physical consistency remain key limitations, most frameworks depend on textual priors rather than explicit dynamics.
Critical constraints include enforcing speed and acceleration bounds, collision avoidance, and map compliance (lane adherence, right-of-way).
A recurring trade-off arises between controllability and fidelity: linguistic prompts enable intuitive design but complicate physically plausible synthesis and balanced coverage.
To assess practical utility, we suggest reporting synthetic-to-real transfer deltas (ADE/FDE) and scenario-coverage metrics over taxonomy categories.
Future work may integrate physics-aware constraints, human-in-the-loop validation, and standardized benchmarks to evaluate realism, diversity, and generalization of language-generated data.


\section{Language-Based Reasoning and Interpretability for Trajectory Prediction}\label{sec:taxonomy_5}
As trajectory predictors become increasingly complex, understanding how and why they produce specific outputs is essential for safety, trust, and effective human–AI collaboration.
Recent studies integrate LLMs not only to generate trajectories but also to explain and reason about them—linking motion forecasting with natural language reasoning.
By articulating goals, intermediate intentions, or decision chains via chain-of-thought (CoT) prompting~\citep{wei2022chain}, these approaches aim to expose the logic behind autonomous decisions.


GPT-Driver~\citep{mao2023gpt} represents one of the first unified frameworks to generate both future maneuvers and textual rationales using GPT-style decoding.
This joint formulation provides interpretable motion planning but struggles with causal grounding, often yielding explanations that sound reasonable yet lack mechanistic accuracy.


LC-LLM~\citep{peng2024lcllm} extends this idea with a dual-task setup, predicting lane-change actions and their textual justifications (e.g., ``the vehicle merges left due to a slower car ahead'').
Such narrative explanations improve user trust but risk post-hoc rationalization, where language aligns with outcomes rather than underlying decision logic.
Taking a broader view, xTP-LLM~\citep{guo2024towards} adapts LLaMA2-7B-Chat for traffic flow forecasting, translating multimodal signals, weather, points of interest, and historical flows into structured prompts.
It demonstrates the scalability of LLM reasoning for aggregate prediction, though extending interpretability to agent-level behavior remains unsolved.

More recently, CoT-Drive~\citep{liao2025cot} explicitly treats trajectory prediction as a reasoning task.
Instead of direct regression, the model verbalizes intermediate steps (e.g., ``the car approaches a stop sign, slowing down is expected'') before final trajectory generation.
This structured reasoning improves both accuracy and interpretability but incurs high computational cost and limited quantitative evaluability, as reasoning traces often diverge from ground-truth dynamics.

Collectively, these works shift language from a descriptive layer to a reasoning mechanism, using verbalized logic to reveal model intent and decision processes.
Through waypoint explanation, rationale generation, and CoT-based decision chains, LLMs are emerging as a bridge between predictive performance and human-centered transparency.


\noindent\textbf{Discussion.}
Language-based reasoning offers a path toward interpretable and trustworthy motion forecasting by verbalizing internal decision logic.
Yet current explanations are mostly descriptive rather than causal—LLMs often produce plausible narratives without faithfully mirroring internal computations.
Evaluating faithfulness remains difficult, practical probes include counterfactual sensitivity (does the rationale change under scene edits?) and rationale–action consistency (does removing rationale tokens affect predicted motion?).
A persistent tension arises between interpretability and reliability: detailed rationales improve human understanding but risk inconsistency or ambiguity.
Future work should align textual rationales with latent decision representations, develop faithfulness metrics, and explore causal reasoning modules that ensure both correctness and clarity.
A lightweight direction is to distill CoT traces into compact latent rationales, preserving decision-critical cues while mitigating decoding cost.


%% file: section/5_challenges.tex
\section{Challenges and Future Directions}\label{sec.challenges}
Despite remarkable progress, LLM-based trajectory prediction still faces intertwined challenges across representation, reasoning, and evaluation.
These are not isolated technical bottlenecks but manifestations of a deeper tension between linguistic abstraction and spatial grounding.
Below, we summarize key challenges and outline promising directions toward more scalable, interpretable, and physically consistent predictors.


\noindent\textbf{Tokenization and Representation.}
A core challenge lies in converting continuous, multi-agent trajectories into discrete sequences suitable for LLM processing.
Naïve numeric tokenization often leads to bloated or lossy representations.
Recent work such as~\citep{philion2023trajeglish} discretizes vehicle motion into compact token vocabularies while preserving centimeter-level accuracy, enabling realistic multi-agent simulation via next-token prediction.
However, such token schemes are typically hand-crafted and may not generalize across agents or domains.
Future directions include learned or hierarchical tokenizers, semantic abstractions (e.g., maneuver- or intent-level tokens), and hybrid representations that couple discrete language tokens with continuous latent embeddings.
The central difficulty remains balancing spatial fidelity with linguistic tractability, a fundamental trade-off that defines this research frontier.


\noindent\textbf{Prompt Design and Alignment.}
Effectively steering LLMs for trajectory reasoning depends on prompts that encode scene semantics, agent states, and prediction objectives.
Despite early success, prompt alignment remains fragile and often task-specific.
Scalable strategies such as soft prompting, instruction tuning, or reinforcement learning from feedback (RLHF/RLAIF) may enhance cross-domain generalization and adaptive control.
Ultimately, progress hinges on moving beyond handcrafted templates toward self-aligned interfaces, where models autonomously interpret and refine instructions based on context and feedback.


\noindent\textbf{Reasoning and Prediction Coherence.}
LLMs offer an opportunity to embed commonsense and causal reasoning into motion forecasting, yet their temporal and physical reasoning remain limited.
While \citet{peng2024lcllm} demonstrates the benefit of chain-of-thought reasoning for explaining long-term forecasts, most models still produce inconsistent or physically implausible outputs~\citep{munir2024exploring}.
Enhancing reasoning may require physics-informed priors, structured scene graphs, or verifiable planning modules that constrain generation.
Developing reasoning-aligned datasets with explicit causal annotations could further encourage robust decision-making.
The broader challenge is to infuse linguistic reasoning with causal, physically grounded logic, while maintaining the flexibility that makes LLMs powerful.


\noindent\textbf{Multimodal Context Integration.}
Real-world trajectory forecasting demands understanding of visual perception, map topology, and language instructions.
Several works address this by projecting scene features into token-compatible embeddings~\citep{felemban2024imotion} or encoding environmental cues as text~\citep{chen2024driving}.
However, coupling perception with prediction remains difficult due to modality mismatches and limited context windows.
Emerging architectures that unify vision and language modeling (e.g., VLMs, multimodal Transformers) or use graph-enhanced scene prompts show promise for integrating heterogeneous signals.
The ultimate goal extends beyond simple fusion, it is representational unification, where vision, map, and language share a coherent reasoning space.


\noindent\textbf{Transparency and Interpretability.}
One of the most promising byproducts of incorporating LLMs is their ability to generate natural language rationales for predicted behaviors.
Models such as LC-LLM~\citep{peng2024lcllm} and xTP-LLM~\citep{guo2024towards} enhance human trust by coupling predictions with textual explanations.
However, ensuring fidelity, that explanations reflect actual decision processes, remains challenging.
Future research may integrate internal alignment losses, counterfactual reasoning, or multi-task learning, where explanation quality directly influences prediction.
A persistent tension exists between interpretability and autonomy: systems that explain well may not always act optimally.
Achieving causal, faithful, and efficient interpretability frameworks will be crucial for safety-critical deployment.


\noindent\textbf{Evaluation and Benchmarking.}
As LLM-based predictors generate both trajectories and rationales, traditional metrics such as ADE, FDE, and MR (see Appendix~\ref{sec:appendix_A}) are no longer sufficient.
Existing benchmarks rarely assess multi-agent coordination, reasoning coherence, or language-grounded objectives.
Moreover, textual outputs lack standardized evaluation protocols.
Future benchmarks should jointly evaluate trajectory accuracy, semantic plausibility, and explanation faithfulness, supported by language-augmented datasets with aligned ground truth.
Next-generation evaluations must capture holistic decision quality, covering reasoning correctness, behavioral safety, and linguistic reliability, rather than isolated positional errors.


%% file: section/7_conclusion.tex
\section{Conclusion and Discussion}
In this survey, we reviewed the growing research at the intersection of language modeling and trajectory prediction. 
By organizing recent work into five categories, we highlight how LLMs and natural language representations can be used to enhance prediction performance, enable data generation, improve interpretability, and support multimodal reasoning. 
This trend indicates a shift from traditional numeric or vision-only models toward more semantically enriched and cognitively aligned systems. 
While the language integration offers promise in generalization, transparency, and user interaction, it also introduces new challenges in grounding and evaluation.
We hope this survey provides a useful foundation for advancing research at the intersection of language and trajectory prediction.

\noindent \textbf{AI Disclosure.}
We used ChatGPT solely for grammar correction and language polishing. All research content, literature analysis, and writing were conducted independently by the authors.

\noindent \textbf{Potential Risks.}
We note that emphasizing LLM-based approaches could amplify existing data or representation biases. 
We encourage readers to interpret the reviewed works critically and consider their broader implications before deploying them in real-world settings.


%% file: section/limitations.tex
\section*{Limitations}
Given the rapid development of natural language processing and trajectory prediction, it is possible that some recent approaches may not have been included in this survey. Nonetheless, we have made every effort to ensure that the most representative and impactful works are covered.
In Section~\ref{sec:taxonomy_1}, we do not attempt to exhaustively review all Transformer-based trajectory prediction models, as Transformer architectures and attention mechanisms have become foundational components in the field. 
Instead, we selectively highlight well-recognized, open-source, and influential works that best illustrate the evolution of language modeling paradigms in trajectory prediction.
Finally, some references could reasonably belong to multiple taxonomic categories. In such cases, we categorize each work according to what we consider to be its primary contribution, while acknowledging that many methods operate at the intersection of several themes.

%% file: section/ethics.tex
\section*{Ethics Statement}
We confirm that this survey adheres to ethical research standards. All analyzed works are publicly available, and we do not involve any human subjects or personally identifiable data. 
The purpose of this survey is to facilitate academic understanding and responsible development of language-based trajectory prediction. 
We have acknowledged prior work accurately and its original contributions.

%% file: section/x_appendix.tex
\section*{\centering Appendix}
\section{Background on Trajectory Prediction}\label{sec:appendix_A}
Trajectory prediction seeks to forecast the future positions 
\(\{\mathbf{p}_{t,i}\}_{t=1}^T \subset \mathbb{R}^2\) of \(N\) agents (indexed by \(i=1,\dots,N\)), 
given their past positions \(\{\mathbf{p}_{t,i}\}_{t=-\tau+1}^0\).

\noindent\textbf{Problem Formulation.}  We denote the observed history of length \(\tau\) for agent \(i\) as
\[
  \mathbf{X}_{{\rm hist},i}
  = (\mathbf{p}_{-\tau+1,i}, \dots, \mathbf{p}_{0,i}),
\]
and the target future as
\[
  \mathbf{Y}_{i}
  = (\mathbf{p}_{1,i}, \dots, \mathbf{p}_{T,i}).
\]
Most modern methods learn a conditional distribution
\[
  p\bigl(\mathbf{Y}_{i} \,\big\vert\, \mathbf{X}_{{\rm hist},i},\,\mathcal{C}_{i}\bigr),
\]
where \(\mathcal{C}_{i}\) may include scene context (HD-maps, images), 
social interaction graphs, or goal priors.

\noindent\textbf{Prediction Outputs.}  For each test sample \(i\), the model typically
generates \(K\) hypotheses
\(\{\hat{\mathbf{p}}^{(k)}_{1:T,i}\}_{k=1}^K\).

\noindent\textbf{Metrics.}  
Let \(\mathbf{p}_{1:T,i} = (\mathbf{p}_{1,i},\dots,\mathbf{p}_{T,i})\) denote the ground-truth, and \(\hat{\mathbf{p}}^{(k)}_{1:T,i}\) the \(k\)-th predicted trajectory. 
Then:
\begin{itemize}[leftmargin=*, labelsep=0.5em]
  \item \textbf{Average Displacement Error (ADE)} for sample \(i\), hypothesis \(k\):
    \[
      \mathrm{ADE}_{i}^{(k)}
      = \frac{1}{T}\sum_{t=1}^{T}
        \bigl\lVert \hat{\mathbf{p}}^{(k)}_{t,i} - \mathbf{p}_{t,i} \bigr\rVert_2.
    \]

  \item \textbf{Final Displacement Error (FDE)} for sample \(i\), hypothesis \(k\):
    \[
      \mathrm{FDE}_{i}^{(k)}
      = \bigl\lVert \hat{\mathbf{p}}^{(k)}_{T,i} - \mathbf{p}_{T,i} \bigr\rVert_2.
    \]

  \item \textbf{Minimum ADE (minADE)} for sample \(i\):
    \[
      \mathrm{minADE}_{i}
      = \min_{k=1,\dots,K}\;\mathrm{ADE}_{i}^{(k)}.
    \]

  \item \textbf{Minimum FDE (minFDE)} for sample \(i\):
    \[
      \mathrm{minFDE}_{i}
      = \min_{k=1,\dots,K}\;\mathrm{FDE}_{i}^{(k)}.
    \]

  \item \textbf{Miss Rate (MR)} at threshold \(\delta\):
    \[
      \mathrm{MR}(\delta)
      = \frac{1}{N}\sum_{i=1}^N
        \mathbf{1}\!\Bigl[
          \min_{k=1,\dots,K}
            \bigl\lVert \hat{\mathbf{p}}^{(k)}_{T,i} - \mathbf{p}_{T,i}\bigr\rVert_2
          > \delta
        \Bigr],
    \]
    where \(\mathbf{1}[\cdot]\) is the indicator function.
\end{itemize}

\noindent\textbf{Benchmarks and Datasets.}
Commonly used publicly available benchmarks include:
\begin{itemize}[leftmargin=*, labelsep=0.5em]
  \item \textbf{ETH/UCY}~\citep{pellegrini2009You, lerner2007crowds}.
  \item \textbf{SDD}~\citep{robicquet2016learning}.
  \item \textbf{INTERACTION}~\citep{zhan2019interaction}.
  \item \textbf{nuScenes}~\citep{caesar2020nuscenes}.
  \item \textbf{Argoverse 1 \& 2}~\citep{chang2019argoverse, wilson2023argoverse}.
  \item \textbf{Waymo Open Motion Dataset (WOMO)}~\citep{ettinger2021large}.
  \item \textbf{Sports-Traj}~\citep{xu2025sports}.
\end{itemize}

\section{Survey Methodology}\label{sec:appendix_meth}
To ensure both comprehensive coverage and methodological transparency, we followed a structured process to select and organize relevant literature at the intersection of large language models (LLMs) and trajectory prediction.

\noindent\textbf{Time frame and data sources.}
Our survey focuses on works published between 2022 and 2025, reflecting the rapid development of LLMs in recent years. We sourced papers from arXiv, ACL Anthology, Semantic Scholar, and Google Scholar. These repositories were chosen to cover both peer-reviewed publications and influential preprints.

\noindent\textbf{Search process.}
We used keyword-based queries including \textit{``trajectory prediction''}, \textit{``trajectory forecasting''}, \textit{``motion forecasting''}, and \textit{``large language model''} to identify relevant works. Boolean combinations of terms (e.g., ``trajectory prediction and LLM'') were used to refine results. We also performed backward and forward citation tracking on several seed papers to ensure broader coverage.

\noindent\textbf{Selection criteria.}
We included papers that use language models, either through prompting, fine-tuning, or alignment, to guide, improve, or explain trajectory prediction tasks. Both agent-level motion forecasting (e.g., vehicles, pedestrians) and simulation-oriented prediction were considered. We excluded studies that focus solely on traffic flow modeling, origin-destination estimation, or aggregate-level forecasting without agent-specific predictions.

\noindent\textbf{Scope.}
After initial screening, we reviewed over 70 papers and retained 35 representative works for in-depth analysis. These were selected based on relevance, technical novelty, diversity of modeling approaches, and influence (e.g., venue, citations, or novelty). Preprints from arXiv were included when they presented novel approaches or had gained significant attention within the community.

\section{Relation with Other Surveys}\label{sec:appendix_B}
Several recent surveys have explored the integration of large language models (LLMs) and vision-language models (VLMs) into autonomous driving systems. These efforts primarily focus on perception, planning, and decision-making, highlighting applications such as multimodal scene understanding~\citep{zhou2024vision}, instruction following~\citep{cui2024survey}, and end-to-end control stacks~\citep{yang2023llm4drive}. However, trajectory prediction, a core module in the autonomy pipeline, has received relatively limited attention within this emerging paradigm.

In contrast, our survey provides the first comprehensive review dedicated to LLM-based trajectory prediction, focusing on how language modeling, prompting, data generation, and interpretability mechanisms are reshaping this subfield. We emphasize the use of LLMs beyond high-level planning or user interaction, instead targeting their application in modeling, guiding, or generating fine-grained future trajectories. To our knowledge, this is the first taxonomy to structure these advances explicitly around language as a modeling axis.

We also differentiate our work from traditional trajectory prediction surveys~\citep{huang2022survey, madjid2025trajectory}, which extensively cover geometric, learning-based, and multimodal methods, but largely overlook the recent surge of language-driven approaches. Our survey fills this gap by highlighting how natural language is used as both input and supervision to enhance prediction.
Finally, one recent survey on knowledge integration in prediction and planning~\citep{manas2025knowledge} summarizes rule-based and symbolic reasoning frameworks, they do not systematically cover the use of pretrained LLMs or generative prompting techniques. Our work complements theirs by showcasing how foundation models introduce new forms of semantic grounding, controllability, and reasoning into the trajectory prediction.

\section{Comparative Overview of Language-Based Methods}\label{sec:appendix_C}

\input{table/method_2}
\input{table/method_3}
\input{table/method_4}
\input{table/method_5}
To provide a clearer comparison of existing work, we present a summary covering the four categories that explicitly utilize language models: 
(1) Direct Trajectory Prediction with Language-Based Predictors, (2) Language-Guided Scene Understanding for Trajectory Prediction
, 
(3) Language-Driven Data Generation for Trajectory Prediction, and 
(4) Language-Based Reasoning and Interpretability for Trajectory Prediction. These categories are summarized in Table~\ref{tab:method_2}, Table~\ref{tab:method_3}, Table~\ref{tab:method_4}, and Table~\ref{tab:method_5}, respectively, highlighting how different methods incorporate LLMs, their prompt design, fine-tuning strategies, and datasets.
We do not include a table for the first taxonomy, \textit{Trajectory Prediction via Language Modeling Paradigms}, as the methods in this category do not explicitly employ language models, but rather draw structural inspiration from language modeling paradigms.

%% file: table/method_2.tex
\begin{table*}
\centering
\scalebox{0.75}{
\begin{tabular}{lcp{4cm}p{4cm}p{4cm}p{2.5cm}}
\hline
Method & Year & LLM Usage & Prompt Design 
& Fine-tuning Strategy & Dataset  \\
\hline
LMTraj~\citeyearpar{bae2024can} & 2024
& GPT-style Seq2Seq LMs (e.g., T5) with a custom numerical tokenizer and QA-style sentence generation
& QA format with image caption and social questions
& Fully supervised fine-tuning QA objective
& ETH/UCY, SDD
\\
\hline
Exploring~\citeyearpar{munir2024exploring} & 2024
& GPT-2, LLaMA, Zephyr, LLM predicts future trajectory tokens
& GPT-Driver style prompt with structured tokens
& PEFT (LoRA, P-Tuning)
& nuScenes
\\
\hline
iMotion-LLM~\citeyearpar{felemban2024imotion} & 2024
& LLM (GPT-like) used to interpret scene embeddings and human instructions for generating trajectories     
& Natural language instructions + scene embedding tokens      
& LoRA fine-tuning on LLM projection and decoding layers 
& InstructWaymo (Waymo-augmented)
\\
\hline
Traj-LLM~\citeyearpar{lan2024traj} & 2024
& Frozen LLM (GPT-2, LLaMA), used for joint scene-agent representation
& No handcrafted prompt, implicit in tokenization
& Frozen LLM + learnable modules
& nuScenes
\\
\hline
FlightLLM~\citeyearpar{luo2025large} & 2025
& LLaMA 3.1, Mistral, etc., LLM outputs next-step tokens
& Structured prompt of flight context
& PEFT (LoRA, QLoRA, Adapter)
& ADS-B (OpenSky)
\\
\hline
\end{tabular}
}
\caption{Methods for direct trajectory prediction using pretrained language models, highlighting how LLMs are integrated into the prediction pipeline.}
\label{tab:method_2}
\end{table*}

%% file: table/method_3.tex
\begin{table*}
\centering
\scalebox{0.75}{
\begin{tabular}{lcp{4cm}p{4cm}p{4cm}p{2.5cm}}
\hline
Method & Year & LLM Usage & Prompt Design 
& Fine-tuning Strategy & Dataset  \\
\hline
LingRep~\citeyearpar{kuo2022trajectory} & 2022
& LLM (e.g., GPT2) generates intermediate linguistic descriptions, used as latent features for prediction 
& Generates short descriptions from heuristic templates
& LLM is co-trained with the trajectory decoder    
& Argoverse 1
\\
\hline
TextCover~\citeyearpar{keysan2023can}  & 2023
& Uses DistilBERT as text encoder for traffic scene representations
& Structured text prompt describing target agent, past positions, and lanes
& DistilBERT fine-tuned, joint encoder uses frozen weights	
& nuScenes
\\
\hline
CTG++~\citeyearpar{zhong2023language}  & 2023
& Uses GPT-4 to generate differentiable loss functions from language queries	
& Few-shot prompting with helper functions and paired query-loss examples	
& LLM is frozen, only the diffusion model is trained	
& nuScenes
\\
\hline
LLM-Powered~\citeyearpar{zheng2024large} & 2024
& Use GPT-4-V (vision-language model) to interpret TC-Maps and output traffic context	
& Combined image and structured text prompt, includes intention, affordance, and scenario description	
& LLM frozen, only the motion predictor is trained
& WOMD
\\
\hline
LG-Traj~\citeyearpar{chib2024lg} & 2024
& Uses LLM to generate motion cues from observed trajectories (e.g., linear/curved/stationary)	
& Instruction-style prompt using trajectory coordinates in chat template format	
& LLM is frozen, only the encoder-decoder model is trained	
& ETH/UCY, SDD
\\
\hline
VisionTrap~\citeyearpar{moon2024visiontrap} & 2024
& Uses BLIP-2 and LLM to generate and refine scene-level textual descriptions	
& Vision-language captions auto-generated and refined (e.g., ``expected to continue straight'')	
&  LLM/VLM frozen, only the trajectory predictor is trained 
& nuScenes-Text
\\
\hline
\end{tabular}
}
\caption{Methods leveraging large language models to enhance scene and context understanding for trajectory prediction.}
\label{tab:method_3}
\end{table*}

%% file: table/method_4.tex
\begin{table*}
\centering
\scalebox{0.75}{
\begin{tabular}{lcp{4cm}p{4cm}p{4cm}p{2.5cm}}
\hline
Method & Year & LLM Usage & Prompt Design 
& Fine-tuning Strategy & Dataset  \\
\hline
LCTGen~\citeyearpar{tan2023language}
& 2023
& Uses GPT-4 in the Interpreter to convert natural language into structured representations	
& In-context learning and Chain-of-Thought prompting, YAML-like structure + map/actor specification	
& LLM is frozen; only the generator module is trained
WOMO
\\
\hline
SeGPT~\citeyearpar{li2024chatgpt}
& 2024
& Uses ChatGPT (GPT-4) to generate complex and diverse trajectory prediction scenarios	
& Prompt engineering includes persona-setting, reference text, task decomposition, CoT, and zero-shot CoT strategies	
& ChatGPT frozen, no fine-tuning	
& INTERACTION, and SeGPT-generated dataset
\\
\hline
InteractTraj~\citeyearpar{xia2024language}
& 2024
& Uses GPT-4 to convert language descriptions into structured interaction, vehicle, and map codes	
& Multi-part structured prompt (interaction/vehicle/map) with detailed rules and examples	
& LLM is frozen, only the code-to-trajectory decoder is trained	
& WOMO, nuPlan~\citeyearpar{caesar2021nuplan}
\\
\hline
Trajectory-LLM~\citeyearpar{yang2025trajectory}
& 2025
& Uses LLaMA-7B for two-stage ``interaction - behavior - trajectory'' generation	
& Generate behavior text from interaction descriptions, and then translate behavior text into trajectories	
& LLM is frozen, trainable regression heads on top	
& L2T, WOMD, Argoverse
\\
\hline
\end{tabular}
}
\caption{Methods that use large language models to generate synthetic data or augment training sets for trajectory prediction.}
\label{tab:method_4}
\end{table*}

%% file: table/method_5.tex
\begin{table*}
\centering
\scalebox{0.75}{
\begin{tabular}{lcp{4cm}p{4cm}p{4cm}p{2.5cm}}
\hline
Method & Year & LLM Usage & Prompt Design 
& Fine-tuning Strategy & Dataset  \\
\hline
GPT-Driver~\citeyearpar{mao2023gpt}
& 2023
& Uses GPT-3.5 for motion planning as a language modeling task	
& Parameterized perception, ego-state, and mission goal, and prompt with explicit chain-of-thought reasoning	
& Prompting - reasoning - fine-tuning strategy
& nuScenes
\\
\hline
LC-LLM~\citeyearpar{peng2024lcllm} 
& 2024
& Reformulates lane change prediction as language modeling using LLaMA-2-13B	
& Natural language prompt describing vehicle/map context, and CoT-style explanatory reasoning	
& Supervised fine-tuning with LoRA on LLaMA-2	
& highD~\citeyearpar{krajewski2018highd}
\\
\hline

xTP-LLM~\citeyearpar{guo2024towards} 
& 2024
& Uses LLaMA2-7B-chat and LoRA fine-tuned for prediction and explanation generation	
& Multi-modal prompt with spatial, temporal (historical traffic), and external factors, and CoT examples	
& LoRA-based supervised fine-tuning	
& CATraffic (based on LargeST~\citeyearpar{liu2023largest})
\\
\hline

CoT-Drive~\citeyearpar{liao2025cot}
& 2025
& Uses GPT-4 Turbo as teacher, distilled into lightweight edge LMs (e.g., Qwen-1.5, Phi-1.5, TinyLLaMA)	
& 4-step CoT prompt: Background - Interaction - Risk Assessment - Prediction, structured like human reasoning	
& Two-stage: (1) LLM distillation on CoT-labeled text; (2) supervised trajectory prediction	
& NGSIM~\citeyearpar{deo2018convolutional}, highD~\citeyearpar{krajewski2018highd}, MoCAD~\citeyearpar{liao2024bat}, ApolloScape~\citeyearpar{huang2018apolloscape}, nuScenes
\\
\hline
\end{tabular}
}
\caption{Methods focusing on reasoning and interpretability in trajectory prediction via language-based explanations and LLM-generated semantics.}
\label{tab:method_5}
\end{table*}